\pdfoutput=1
\documentclass[journal,twoside,web]{ieeecolor}
\usepackage{generic}
\usepackage{cite}
\usepackage{amsmath,amssymb,amsfonts}
\usepackage{algorithmic}
\usepackage{graphicx}
\usepackage{algorithm,algorithmic}
\usepackage{hyperref}
\hypersetup{hidelinks=true,
  pdftitle={QCAP-VQA: Question-Conditioned Capped Prior-Mass Supervision for Medical Visual Question Answering},
  pdfauthor={Yining Zhao, Catarina Moreira, Mukesh Prasad, Ali Braytee}}
\usepackage{textcomp}
\usepackage{booktabs}
\def\BibTeX{{\rm B\kern-.05em{\sc i\kern-.025em b}\kern-.08em
    T\kern-.1667em\lower.7ex\hbox{E}\kern-.125emX}}
\begin{document}
\title{QCAP-VQA: Question-Conditioned Capped Prior-Mass Supervision for Medical Visual Question Answering}
\author{Yining Zhao, Catarina Moreira, Mukesh Prasad, and Ali Braytee
\thanks{This work has been submitted to the IEEE for possible publication.
Copyright may be transferred without notice, after which this version may
no longer be accessible.}
\thanks{(Corresponding author: Ali Braytee.)}
\thanks{The authors are with the School of Computer Science,
University of Technology Sydney, Ultimo, NSW 2007, Australia
(e-mail: yining.zhao@student.uts.edu.au; catarina.moreira@uts.edu.au;
mukesh.prasad@uts.edu.au; ali.braytee@uts.edu.au).}}

\maketitle

\begin{abstract}
Medical visual question answering models can answer correctly
without attending to the anatomy that the question names. Automatic region
priors allow spatial supervision without expert masks, but they are noisy,
and it is unclear how strongly they should shape attention and how their
effect should be verified. We present QCAP-VQA, which supervises a subset
of cross-attention heads with a question-conditioned prior built from
ranked segmentation proposals. Its capped prior-mass objective penalizes
insufficient attention inside the prior only below a threshold. We
evaluate attention against SLAKE structure boxes that play no role in
training, separating attention on the named anatomy from how attention
shifts between questions about the same image. QCAP-VQA matches the
accuracy of a baseline without spatial supervision (84.26\% vs.\
84.23\%), and every evaluated prior-supervised model places more attention
on the annotated structures. Over a 64-fold range of regularization
strength, the capped objective keeps prior mass between 0.67 and 0.77,
whereas a pooled region-mass objective drives it to 0.97 and loses
accuracy. The cost lies in question conditioning: as the constraint widens
or strengthens, attention shifts agree less with the expert-defined
contrast. Across 13 configurations, prior-referenced and expert-referenced
evaluation rank the models in largely opposite order (Spearman
correlation of $-0.87$), and the capped models it favors track the
prior's own errors. Because clinical use of medical visual question
answering requires answers grounded in the queried anatomy, supervision
from automatic priors should be judged against independent anatomical
references.
\end{abstract}


\section{Introduction}
\label{sec:introduction}

\IEEEPARstart{M}{edical} visual question answering (Med-VQA) asks a model
to answer a clinical question about a medical image. Most Med-VQA methods
raise answer accuracy through medical vision--language pretraining and
multimodal fusion~\cite{eslami2023pubmedclip, chen2022multi,
lin2023pmcclip, gu2024lapa}, but accuracy cannot show whether an answer
depends on the structure that the question names; a model can also exploit
regularities in the question text or unrelated image features. Attention
maps do not explain an answer~\cite{jain2019attention,
wiegreffe2019attention}, but they are one of the few internal quantities
that can be inspected against anatomy and shaped during training.

Shaping attention requires a target. Attention or explanation supervision
in general-domain visual question answering uses human attention maps or
annotated evidence regions~\cite{selvaraju2019taking,ross2017right}, and
region-supervised Med-VQA likewise relies on annotated regions of
interest~\cite{chen2025r}. Such annotation is costly and rarely available
at the scale of a VQA training set. A$^3$Tune~\cite{chang2025focus}
removes this requirement for medical vision--language models: it builds
weak labels from zero-shot segmentation proposals, refines them with
BiomedCLIP~\cite{zhang2023biomedclip} into prompt-aware targets, and tunes
selected visually critical attention heads toward them. Automatic priors
and head selection are therefore established. What remains open is how
such a prior should be used once it is known to be noisy: whether
supervision should keep pushing attention toward the prior or stop once a
minimum share of attention lies inside it, and how the effect should be
verified, since a score computed against the same prior can only report
agreement with it. Grounded Med-VQA benchmarks such as HEAL-MedVQA~\cite{nguyen2025localizing}
evaluate localization against physician-annotated masks, but whether
agreement with an automatic training prior predicts agreement with such
references has, to our knowledge, not been examined.

We study both questions with QCAP-VQA. Like A$^3$Tune, it builds its
prior automatically, here from MedSAM~\cite{ma2024segment} proposals
ranked by BiomedCLIP, and supervises a subset of $K$ of eight
cross-attention heads; it differs in how the prior enters the loss and in
how the effect is verified. The form of the loss sets how much
of a noisy prior the model must reproduce. Squared-error (MSE) matching has
a single minimizer, uniform attention inside the prior, and a pooled
region-mass objective adapted from A$^3$Tune keeps rewarding mass until
attention is almost confined to the prior. Our capped objective only
requires each constrained head to place a fraction $\tau$ of its attention
inside the prior and applies no spatial pressure beyond it. To verify the
effect, we score attention against the organ and lesion boxes released
with SLAKE~\cite{liu2021slake}, which play no role in training or
checkpoint selection. We measure how much attention falls on the structure
a question names and how attention shifts between two questions about the
same image, against both the expert boxes and the automatic priors, and
compare the two references on the same trained models.

The contributions of this work are as follows.
\begin{itemize}
\item \emph{A threshold-saturated prior-mass objective for noisy
automatic priors.} Each constrained head must place a minimum share of its
attention inside the prior, beyond which no spatial pressure applies.
Unlike objectives that keep rewarding prior mass, it keeps both the
induced prior mass and answer accuracy stable across regularization
strengths.
\item \emph{The anatomical alignment--question conditioning trade-off.}
More attention on the named structure does not imply a correct
question-conditioned shift: as more heads are constrained under the
capped objective, attention on
the structure rises while expert-referenced question conditioning falls
from $+0.88$ to $-1.67$~percentage points, a change that accuracy does not
reveal. Which heads are constrained matters much less than how many.
\item \emph{An expert-referenced test of what supervision transfers.} A
pairing intervention and independent expert boxes show that spatial
supervision transfers the prior's question-specific structure, errors
included, and that prior-referenced and expert-referenced evaluation rank
the same models in largely opposite order. Agreement with the training
prior is therefore not evidence of anatomical grounding.
\end{itemize}

\section{Method}
\label{sec:method}

\begin{figure*}[t]
  \centering
  \includegraphics[width=\textwidth]{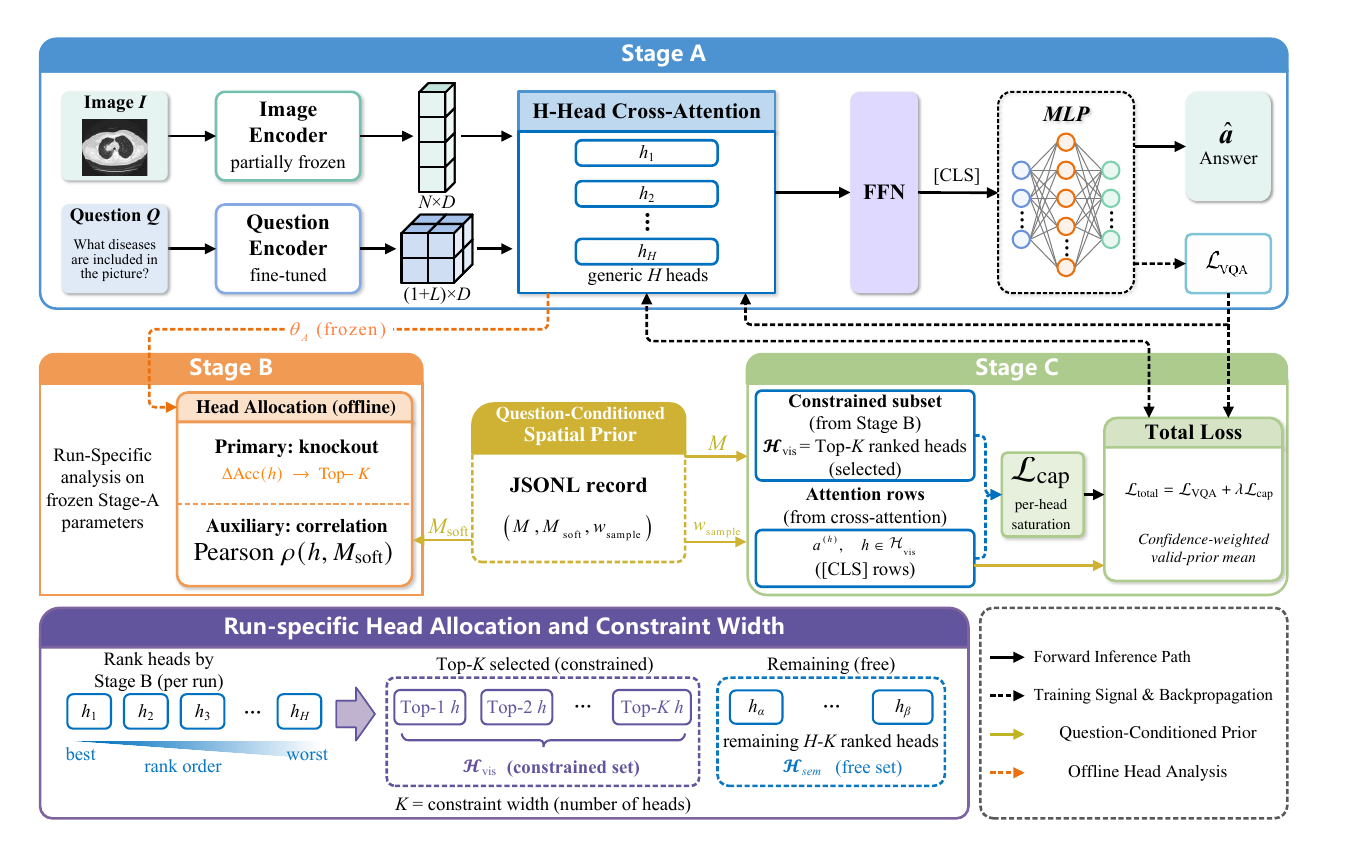}
  \caption{Overview of QCAP-VQA. Stage A trains the VQA model, Stage B
  selects $K$ heads within each run, and Stage C applies capped
  prior-mass supervision to them ($\mathcal{L}_{\mathrm{cap}}$:
  $\mathcal{L}_{\mathrm{att}}$ of Eq.~\eqref{eq:capped_loss}).
  $\mathcal{H}_{\mathrm{vis}}$/$\mathcal{H}_{\mathrm{sem}}$: constrained
  and unconstrained heads, not fixed head functions.}
  \label{fig:framework}
\end{figure*}

\subsection{Framework and Spatial-Prior Input}
\label{sec:method:overview}

QCAP-VQA adds question-conditioned spatial supervision during
training in three stages (Fig.~\ref{fig:framework}) and leaves the
inference architecture unchanged. Expert anatomical annotations are
reserved for independent evaluation and are not used to generate
training priors or optimize the VQA model.

In Stage A, a partially frozen BiomedCLIP
ViT-B/16~\cite{zhang2023biomedclip} encodes image $I$ into
$N=196$ patch tokens of dimension $D=768$, excluding the visual
[CLS] token. A fine-tuned PubMedBERT~\cite{gu2021domain}
encodes question $Q$ into $1+L$ text tokens. After projection
into a common feature dimension, image features
$\mathbf{F}\in\mathbb{R}^{N\times D}$ provide keys and values,
while textual features
$\mathbf{T}\in\mathbb{R}^{(1+L)\times D}$ provide queries to
a single cross-attention layer with $H=8$ heads.
A feed-forward block and classifier predict the answer from
the fused textual [CLS] representation by minimizing the VQA
cross-entropy loss $\mathcal{L}_{\mathrm{VQA}}$.
We denote the resulting Stage-A parameters by $\theta_A$.

For head $h$, let $\mathbf{A}^{(h)}$ denote its cross-attention
matrix and
$\mathbf{a}^{(h)}=\mathbf{A}^{(h)}[0,:]\in\Delta^{N-1}$
the [CLS]-to-image attention distribution, where
$\Delta^{N-1}$ is the probability simplex over visual patches.
All spatial constraints and attention diagnostics operate on
these attention rows.

\paragraph{Question-conditioned spatial priors.}
Spatial priors are generated once, offline, before head allocation
and regularized training. A rule-based parser extracts the target
anatomy, spatial modifiers, and question type, from which it builds
positive text prompts and, for existence and health questions,
negative text prompts. MedSAM~\cite{ma2024segment} produces region
candidates from anatomical-box, adaptive-grid, and automatic-mask
prompts. After intersection-over-union deduplication, each candidate
is scored by its BiomedCLIP similarity to the positive prompts, reduced
by its similarity to the negative prompts and computed over cropped,
padded, and masked views, together with mask quality, anatomical
position, and, for comparison questions, region area. The weights
depend on question type, and modality-specific checks, such as the
field of view for CT, remove implausible candidates.

For training example $i$, up to $R=3$ top-ranked candidate boxes are
projected onto the $14\times14$ visual patch grid.
Their union defines the binary spatial prior
$M^{(i)}\in\{0,1\}^{N}$.
A complementary soft prior $M_{\mathrm{soft}}^{(i)}$ is obtained
by combining box-scaled Gaussian maps through element-wise
maximum, with standard deviations equal to half the corresponding
box width and height. The binary union is used for spatial
supervision, whereas the soft prior supports auxiliary
head-correlation analysis.

The composite score of the highest-ranked candidate defines a sample
confidence weight $w_{\mathrm{sample}}^{(i)}$.
Each image--question pair provides
$(M^{(i)},M_{\mathrm{soft}}^{(i)},w_{\mathrm{sample}}^{(i)})$.
Because proposal ranking depends on the question, different
questions about the same image may produce different priors.
Expert anatomical boxes are never inputs to this procedure.

\subsection{Stage B: Run-Specific Head Allocation}
\label{sec:method:stageb}

Stage B determines which attention heads receive spatial
supervision before Stage C begins. It operates on the frozen
Stage-A checkpoint $\theta_A$ and uses a deterministic validation
subset selected by question wording, including organ identification,
localization, and laterality queries. The same selection rule
is applied across training seeds. The primary allocation criterion
uses neither test questions, expert annotations, nor spatial priors.

Following attention-head intervention
studies~\cite{voita2019analyzing,michel2019sixteen},
we perform a [CLS]-row knockout analysis. For each head $h$,
its [CLS] attention row is replaced by a uniform distribution
during validation inference, without updating model parameters.
The resulting accuracy reduction is

\begin{equation}
  \Delta\mathrm{Acc}(h)
  =
  \mathrm{Acc}(\theta_A)
  -
  \mathrm{Acc}(\theta_A\setminus h),
  \label{eq:knockout}
\end{equation}

where $\theta_A\setminus h$ denotes this forward-pass
intervention rather than removal or retraining of the head.
The $H$ heads are ranked in descending order of
$\Delta\mathrm{Acc}(h)$, requiring $H+1$ validation evaluations.

The top $K$ heads form the constrained set
$\mathcal{H}_{\mathrm{vis}}$, while the remaining heads
constitute $\mathcal{H}_{\mathrm{sem}}$.
These labels describe how the spatial loss is assigned,
rather than fixed visual or semantic roles across
model initializations.

We additionally examine correlation-based allocation using
the validation-mean Pearson correlation between
$\mathbf{a}^{(h,i)}$ and $M_{\mathrm{soft}}^{(i)}$,
together with random head allocation as an ablation.
The selected subset is recomputed for each Stage-A run
and held fixed during Stage C.

The principal allocation variable is the constraint width
$K=|\mathcal{H}_{\mathrm{vis}}|$.
Because spatial penalties are summed across constrained heads,
increasing $K$ expands the regularized portion of the
attention layer at unchanged per-head coefficient $\lambda$;
it does not redistribute a fixed total regularization budget.

\subsection{Stage C: Capped Prior-Mass Constraint}
\label{sec:method:stagec}

Stage C continues training from the Stage-A weights, applying
spatial supervision only to
$\mathcal{H}_{\mathrm{vis}}$.
For example $i$ and constrained head $h$, the attention mass
assigned to its question-conditioned binary prior is

\begin{equation}
  g_h^{(i)}
  =
  \sum_{n=1}^{N}a_n^{(h,i)}M_n^{(i)},
  \qquad 0\leq g_h^{(i)}\leq1.
  \label{eq:prior_mass}
\end{equation}

\paragraph{Full-map matching.}
To motivate a mass-based constraint, consider squared-error
matching between an attention distribution
$\mathbf{a}\in\Delta^{N-1}$ and a nonempty binary prior $M$
occupying $r=\sum_n M_n\geq1$ patches, with $g=\mathbf{a}^{\top}M$:

\begin{equation}
\begin{aligned}
  \mathcal{L}_{\mathrm{MSE}}(\mathbf{a},M)
  &=
  \frac{1}{N}\sum_{n=1}^{N}(a_n-M_n)^2\\
  &=
  \frac{\|\mathbf{a}\|_2^2-2g+r}{N}.
\end{aligned}
\label{eq:mask_mse}
\end{equation}

Its unique minimizer over the probability simplex assigns
$a_n^\star=1/r$ inside $M$ and zero attention elsewhere,
with minimum $(r-1)^2/(rN)$. Increasing its coefficient
therefore pushes attention toward this single map, even
where the automatic prior is incomplete or inaccurate.

\paragraph{Threshold-saturated prior-mass loss.}
Rather than prescribing the complete attention distribution,
we introduce a spatial penalty that becomes zero once each
constrained head assigns at least a fraction $\tau\in(0,1]$ of its
attention to the prior:

\begin{equation}
  \mathcal{L}_{\mathrm{att}}^{(i)}
  =
  -\sum_{h\in\mathcal{H}_{\mathrm{vis}}}
  \log\!\left[
    \min\!\left(
      \frac{g_h^{(i)}}{\tau},1
    \right)
  \right].
  \label{eq:capped_loss}
\end{equation}

For $0<g_h^{(i)}<\tau$, the derivative of the individual
penalty with respect to region mass is $-1/g_h^{(i)}$ (apart from
a small numerical floor that prevents $\log 0$). The penalty becomes zero at
$g_h^{(i)}=\tau$ and has zero spatial gradient strictly above
this threshold.

Each constrained head admits the zero-loss
feasible set

\[
  \left\{
  \mathbf{a}\in\Delta^{N-1}:
  \mathbf{a}^{\top}M\geq\tau
  \right\},
\]

rather than a unique target attention distribution.
Once the threshold is satisfied, the VQA objective can
shape the remaining attention allocation without continued
spatial pressure to concentrate on the prior.
An uncapped $-\log g$ objective, by contrast, keeps
rewarding region mass up to one. The threshold $\tau$ is
the saturation point of the spatial loss, not an upper
bound on the attention mass of the trained model.

\paragraph{Pooled region-mass comparison.}
For an objective-level comparison, we adapt the pooled
region-mass objective of A$^3$Tune~\cite{chang2025focus}
to the same binary union prior.
It first averages the prior mass across constrained heads,

\[
  \bar g^{(i)}
  =
  \frac{1}{K}
  \sum_{h\in\mathcal{H}_{\mathrm{vis}}}g_h^{(i)},
\]

and then applies the squared residual

\begin{equation}
  \mathcal{L}_{\mathrm{pooled}}^{(i)}
  =
  \left(1-\bar g^{(i)}\right)^2.
  \label{eq:pooled_loss}
\end{equation}

Unlike the per-head capped loss, it keeps rewarding the
average prior mass until every constrained head attends
only to the prior. We compare the objectives in prior-mass
concentration, answer accuracy, and question-conditioned
attention.

\subsection{Training Objective and Inference}
\label{sec:method:training}

For a mini-batch $\mathcal{B}$, let

\[
  \mathcal{B}'
  =
  \left\{
  i\in\mathcal{B}:
  w_{\mathrm{sample}}^{(i)}>0,\
  \sum_n M_n^{(i)}>0
  \right\}
\]

denote the samples with valid spatial priors.
The Stage-C training objective is

\begin{equation}
  \mathcal{L}_{\mathrm{total}}
  =
  \mathcal{L}_{\mathrm{VQA}}
  +
  \lambda
  \frac{
    \displaystyle
    \sum_{i\in\mathcal{B}'}
    w_{\mathrm{sample}}^{(i)}
    \mathcal{L}_{\mathrm{att}}^{(i)}
  }{
    \displaystyle
    \sum_{i\in\mathcal{B}'}
    w_{\mathrm{sample}}^{(i)}
  },
  \label{eq:total_loss}
\end{equation}
where $\mathcal{L}_{\mathrm{VQA}}$ is the mean cross-entropy over
the full mini-batch.

For objective ablations, $\mathcal{L}_{\mathrm{att}}^{(i)}$ in
Eq.~\eqref{eq:total_loss} is replaced by the MSE of
Eq.~\eqref{eq:mask_mse} or the uncapped penalty
$-\log g_h^{(i)}$, both summed over constrained heads, or by the
pooled objective of Eq.~\eqref{eq:pooled_loss}, with the same
sample weighting.

Samples without valid priors do not contribute to the
spatial term, which is zero when $\mathcal{B}'$ is empty.
A full-image prior gives $g_h=1$ for every head and thus
zero capped loss, whereas a prior taken from an unrelated
question can still attract attention without the intended
question--region correspondence.

Only the Stage-C weights are kept for inference; prior
generation, head ranking, and the spatial loss are omitted,
so the architecture and parameter count are those of Stage A.

\subsection{Independent Expert-Referenced Evaluation}
\label{sec:method:expert_eval}

Attention concentrated within an automatically generated
prior does not necessarily indicate correct anatomical
alignment or question-conditioned localization.
We therefore evaluate trained attention distributions
against the organ and lesion bounding boxes released with
SLAKE~\cite{liu2021slake}, which are independent of the
priors used during training.

Each annotated box is projected onto the $14\times14$
patch grid using fractional patch coverage, yielding
a reference $E_q\in[0,1]^N$ for question $q$.
Let the model-level attention distribution be the mean
[CLS] attention over all $H$ heads, including unconstrained
ones,

\[
  \bar{\mathbf a}_q
  =
  \frac{1}{H}
  \sum_{h=1}^{H}\mathbf a_q^{(h)}.
\]

We measure expert-region attention mass as
$100\,\bar{\mathbf a}_q^\top E_q$.
Question-level scores are first averaged within each
image and then across images to obtain
$G_{\mathrm{expert}}$.

\paragraph{Question-conditioned response.}
For two questions $(q_1,q_2)$ concerning different
annotated regions of the same image, let $(R_1,R_2)$
denote the corresponding reference maps.
We define question conditioning as

\begin{equation}
  \mathrm{QC}(R_1,R_2)
  =
  100\,
  (\bar{\mathbf a}_{q_1}
   -\bar{\mathbf a}_{q_2})^\top
  (R_1-R_2).
  \label{eq:qcdid}
\end{equation}

Positive QC indicates that, between the two questions,
attention shifts toward the reference region of each
question; negative QC indicates a shift in the opposite
direction. QC measures a relative shift rather than
absolute localization.
We report \emph{Prior-QC} using the automatically
generated binary priors $R_j=M_{q_j}$, which for test
questions serve only as evaluation references, and
\emph{Expert-QC} using the independent anatomical
references $R_j=E_{q_j}$.
Both metrics are evaluated on the same question pairs.

To account for reference-region size, we additionally
compute area-adjusted Expert-QC using unit-sum
anatomical references:

\begin{equation}
\begin{aligned}
  \mathrm{QC}_{\mathrm{area}}
  =100\,
  &(\bar{\mathbf a}_{q_1}
    -\bar{\mathbf a}_{q_2})^\top\\
  &\left(
    \frac{E_{q_1}}{\sum_n E_{q_1,n}}
    -
    \frac{E_{q_2}}{\sum_n E_{q_2,n}}
  \right).
\end{aligned}
\label{eq:area_adjusted_qc}
\end{equation}

To audit the prior itself, we treat it as an attention
map, $\mathbf p_q=M_q/\sum_n M_{q,n}$, and compute its
expert-referenced QC with $\mathbf p_{q_j}$ in place of
$\bar{\mathbf a}_{q_j}$ in Eq.~\eqref{eq:qcdid}.
Prior precision $\sum_n M_{q,n}E_{q,n}/\sum_n M_{q,n}$
and recall $\sum_n M_{q,n}E_{q,n}/\sum_n E_{q,n}$ are
averaged over the target questions; the chance level
$\sum_n E_{q,n}/N$ of a uniform map serves as a reference
for precision.

\paragraph{Evaluation cohort and statistical analysis.}
We associate test questions with the annotated regions
named in their text using the anatomical labels
provided with the corresponding images.
Questions without an unambiguous question--region
correspondence are excluded, yielding 90 target
questions from 49 distinct images.

For question-conditioning analysis, we construct
15 question pairs spanning 13 images using deterministic
question-template and question-ID rules.
Four pairs are matched-template pairs, in which
the two questions share the same wording apart from
the named anatomical target.
The remaining pairs provide cross-template comparisons.

The evaluation cohort follows from these rules alone
and is reused unchanged across models.
Neither expert annotations nor evaluation scores
are used for model training or checkpoint selection.

For paired metrics, multiple pairs from the same image
are averaged before averaging across images, preventing
images with more eligible pairs from receiving greater
weight. Where confidence intervals are reported,
we use 10,000 image-cluster bootstrap resamples,
retaining all question pairs from each sampled image
together. All compared models use the same resampled
images, preserving paired evaluation. Spearman correlation
compares model rankings under the two references, and
Pearson correlation across pairs compares a model's
Expert-QC with that of the prior.

\section{Experiments}
\label{sec:experiments}

\subsection{Datasets and Experimental Settings}
\label{sec:exp:setup}

We evaluate QCAP-VQA on SLAKE-EN~\cite{liu2021slake}
and VQA-RAD~\cite{lau2018dataset}, reporting overall,
open-ended, and closed-ended answer accuracy.
SLAKE-EN contains 4,919 training, 1,053 validation,
and 1,061 test question--answer pairs. Both datasets
are publicly available and de-identified, so no
additional ethical approval was required.
VQA-RAD is used for the published-method benchmark
comparison with its held-out test split, and 15\% of
its training pairs are held out for checkpoint
selection. The VQA-RAD
configuration uses the same $K=3$, $\tau=0.40$, and
$\lambda=1/2$ as the main SLAKE-EN setting, without
dataset-specific hyperparameter tuning.
All subsequent controlled analyses are conducted on
SLAKE-EN, which provides the anatomical annotations
required for the expert-referenced evaluation in
Section~\ref{sec:method:expert_eval}.

Images are resized to $224\times224$ pixels.
On SLAKE-EN, Stage A is trained for 30 epochs with
AdamW and a batch size of 32. The learning rates are
$3\times10^{-5}$ for newly initialized layers and
$3\times10^{-6}$ for trainable pretrained-encoder
parameters. A cosine learning-rate schedule uses
10\% linear warmup.

Stage C resumes from the corresponding Stage-A
checkpoint and runs for 25 epochs with AdamW at
a learning rate of $10^{-5}$, without an additional
learning-rate scheduler. Both stages use cross-entropy
label smoothing of 0.05 and gradient-norm clipping
at 1.0. The checkpoint with the highest overall
validation accuracy is selected for test evaluation.
The baseline without spatial supervision follows the
same Stage-C schedule with $\lambda=0$.

Unless otherwise noted, capped-objective experiments
use $\tau=0.40$ and $\lambda=1/2$.
The main selective setting constrains $K=3$ heads,
while $K=H=8$ provides the full-width endpoint.
SLAKE-EN results are reported across three independent
training seeds (42, 43, and 44), using the mean and
sample standard deviation where applicable.

Our controlled analyses examine spatial objectives
(capped, MSE, uncapped, and pooled), regularization
strength $\lambda$, constraint width $K$, head
allocation, and priors taken from a different image.
For the objective and regularization-strength
comparisons at $K=H=8$, we use matched Stage-A
initializations, training data, spatial priors,
and Stage-C optimization settings within each seed.

Answer accuracy and all attention measures are
computed on the test set, except for the pairing
intervention in Section~\ref{sec:exp:independent},
which is scored against the prior on the larger
validation split. Training-prior mass is recorded
during Stage-C optimization on the training data.

\subsection{Comparison with State-of-the-Art Methods}
\label{sec:exp:benchmark}

\begin{table}[!t]
  \centering
  \caption{Accuracy (\%) on SLAKE-EN and VQA-RAD; published results follow
their own protocols (PubMedCLIP: best variant per dataset). QCAP-VQA:
three-seed mean on SLAKE-EN, single run on VQA-RAD.}
  \label{tab:main_results}
  \begingroup
  \setlength{\tabcolsep}{3pt}
  \renewcommand{\arraystretch}{1.10}
  \footnotesize
  \begin{tabular}{@{}l*{6}{r}@{}}
    \toprule
    & \multicolumn{3}{c}{SLAKE-EN}
    & \multicolumn{3}{c}{VQA-RAD} \\
    \cmidrule(lr){2-4}\cmidrule(l){5-7}
    Method & Overall & Open & Closed
           & Overall & Open & Closed \\
    \midrule
    M3AE~\cite{chen2022multi}
      & 83.25 & 80.31 & 87.82 & 77.01 & 67.23 & 83.46 \\
    PubMedCLIP~\cite{eslami2023pubmedclip}
      & 80.10 & 78.40 & 82.50 & 72.10 & 60.10 & 80.00 \\
    M2I2~\cite{li2023self}
      & 81.20 & 74.70 & 91.10 & 76.80 & 66.50 & 83.50 \\
    CPCR~\cite{liu2023medical}
      & 81.90 & 80.50 & 84.10 & 72.50 & 60.50 & 80.40 \\
    PMC-CLIP~\cite{lin2023pmcclip}
      & 84.30 & 81.90 & 88.00 & 77.60 & 67.00 & 84.00 \\
    CCIS-MVQA~\cite{cai2024counterfactual}
      & 84.08 & 80.12 & 86.72 & 75.06 & 68.78 & 79.24 \\
    LaPA~\cite{gu2024lapa}
      & 84.73 & 82.17 & 88.70 & 79.38 & 68.72 & 86.40 \\
    \midrule
    \textbf{QCAP-VQA} ($K=3$)
      & 84.26 & 82.38 & 87.18 & 76.50 & 67.60 & 82.35 \\
    \bottomrule
  \end{tabular}
  \endgroup
\end{table}

Table~\ref{tab:main_results} compares QCAP-VQA with
representative state-of-the-art Med-VQA methods.
On SLAKE-EN, QCAP-VQA achieves the highest open-ended
accuracy among the compared methods (82.38\%) and an
overall accuracy of 84.26\%. It exceeds CCIS-MVQA in
open-ended, closed-ended, and overall accuracy, outperforms
M3AE (83.25\%), CPCR, M2I2, and PubMedCLIP, and is within
0.5 percentage points of PMC-CLIP (84.30\%) and LaPA
(84.73\%). On VQA-RAD, its 76.50\% overall accuracy exceeds
CCIS-MVQA, CPCR, and PubMedCLIP and is within
0.6 percentage points of M2I2 (76.80\%) and M3AE (77.01\%).

Within our SLAKE-EN setup, the same backbone trained
without spatial supervision reaches 84.23\% overall
accuracy. The capped $K=3$ configuration
matches this accuracy while adding
question-conditioned spatial supervision.

Cross-study benchmark results provide context for
answer prediction performance, but differences in
architectures, pretraining, and evaluation protocols
prevent them from isolating the effect of spatial
supervision. We investigate that effect through
controlled objective comparisons and independent
expert-referenced analyses on SLAKE-EN.

\subsection{Effect of Threshold Saturation}
\label{sec:exp:objective}

Automatically generated spatial priors may include irrelevant
regions or omit parts of the target anatomy. We therefore
investigate whether spatial supervision should continuously
reward attention mass within the prior or stop imposing
additional pressure once a specified mass threshold is reached.
To compare the objectives under matched conditions, we contrast the proposed
capped loss with an A$^3$Tune-style pooled region-mass
objective~\cite{chang2025focus} at full constraint width
($K=8$). Within each seed, both objectives share the same
Stage-A checkpoint, spatial priors, training data, and Stage-C
optimization schedule. We vary the regularization strength
$\lambda$ from $1/32$ to $2$ for both objectives and extend
the pooled sweep to $8$.

\begin{figure}[!t]
  \centering
  \includegraphics[width=\linewidth]{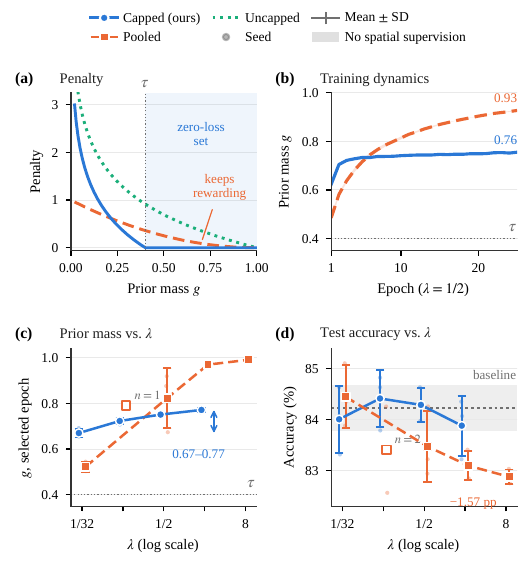}
  \caption{Threshold saturation ($K=8$, $\tau=0.40$). (a)~Illustrative loss
versus prior mass (scales not matched); shaded: capped zero-penalty
region. (b)~Training-prior mass per epoch at $\lambda=1/2$.
(c)~Training-prior mass at the selected epoch. (d)~Test accuracy;
gray: baseline without spatial supervision. Three seeds, except pooled
$\lambda=1/8$ (hollow: two test runs, one training log).}
  \label{fig:objective}
\end{figure}

The objectives differ in their optimization geometry
(Fig.~\ref{fig:objective}(a)). For an individual constrained
head, the capped penalty becomes zero when its prior mass
reaches $g_h=\tau$. It admits a range of
attention distributions with zero spatial penalty, rather
than requiring complete concentration within the prior.
The pooled squared-residual and uncapped log-mass objectives,
in contrast, keep favoring prior mass up to one. This
distinction matters when the spatial prior is imperfect:
additional concentration does not necessarily mean better
anatomical alignment.

The training dynamics follow the same distinction
(Fig.~\ref{fig:objective}(b)). At $\lambda=1/2$, the
capped model reaches an average training-prior mass
of about 0.72 by the third epoch and changes by
only about 0.03 over the remaining 22 epochs, ending at
0.755. The pooled model starts lower, overtakes the
capped model around epoch six, and keeps rising to 0.927
by the final epoch. The capped average stays well above
$\tau=0.40$ because the cap removes the spatial gradient
only for heads and samples that have reached the
threshold; $\tau$ marks where the penalty stops, not a bound
on prior mass.

Across the 64-fold increase of $\lambda$ from $1/32$ to $2$
(Fig.~\ref{fig:objective}(c)), the training-logged prior
mass at the validation-selected epoch rises from 0.670 to
0.771 under capped supervision and from 0.522 to 0.969
under pooled supervision, which reaches 0.990 at
$\lambda=8$. Stronger pooled supervision keeps
concentrating attention on the automatic prior, whereas the
capped mass changes far less once the per-head threshold
is met.

Answer accuracy separates the objectives further
(Fig.~\ref{fig:objective}(d)). Across
$\lambda\in[1/32,2]$, capped accuracy stays between 83.88\%
and 84.42\%, within the seed variation of the baseline
trained without spatial supervision ($84.23\pm0.46\%$).
Pooled accuracy falls from
84.45\% at $\lambda=1/32$ to 82.88\% at $\lambda=8$, and
this 1.57-percentage-point drop appears in each of the
three seeds. Pooled supervision is ahead only at its
weakest setting ($\lambda=1/32$; 0.44 percentage points,
two of three seeds), where it also places the least mass on
the prior. At $\lambda=1/2$ and $2$, capped supervision is
more accurate in all three matched seeds, by 0.82 and 0.79
percentage points on average.

The uncapped log-mass objective removes the threshold
while keeping the per-head formulation, and thus tests its
effect more directly, although these models were trained in
a separate series. At
$\lambda=1/2$ it drives the prior mass to 0.997 ($K=3$) and
0.998 ($K=8$) and lowers accuracy to $83.47\pm0.80\%$ and
$82.81\pm0.71\%$, below the capped models at the same
widths. Without the threshold, attention concentrates almost
entirely on the automatic prior, and answer prediction does
not benefit.

Threshold saturation stops the spatial pressure once
enough prior mass has been assigned. Within the tested
range, capped supervision is less sensitive to
regularization strength in both its induced prior mass
and answer accuracy, whereas the pooled and uncapped
objectives, which keep rewarding region mass, concentrate
attention progressively on the prior and lose accuracy as
they do so.
These results concern optimization and answer accuracy,
not verified anatomical grounding. Whether the resulting
attention lies on the anatomy named in each question, and
whether it changes appropriately between questions, is
examined against expert boxes in
Sections~\ref{sec:exp:tradeoff}
and~\ref{sec:exp:independent}.

\subsection{Anatomical Alignment and Question Conditioning}
\label{sec:exp:tradeoff}

\begin{figure}[!t]
  \centering
  \includegraphics[width=\linewidth]{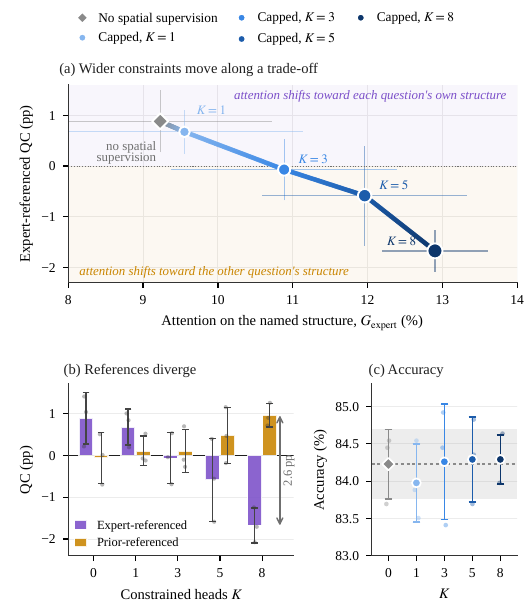}
  \caption{Constraint width $K$ under capped supervision ($\lambda=1/2$; $K=0$:
no spatial supervision). (a)~$G_{\mathrm{expert}}$ versus Expert-QC.
(b)~Expert- and prior-referenced QC. (c)~Test accuracy. Bars: $\pm$ SD
over three seeds; small points: individual seeds.}
  \label{fig:width}
\end{figure}

The previous section examined how strongly the objectives pull attention
toward the prior. Here we measure what this does to attention relative
to the expert boxes, and how the effect depends on how many heads are
constrained, how they are constrained, and which heads are chosen. Two
properties are tracked separately: the attention placed on the structure
named in the question ($G_{\mathrm{expert}}$) and whether attention shifts
between two questions about the same image toward the structures they name
(Expert-QC; negative values indicate a shift in the opposite direction).
Table~\ref{tab:ablation} compares objectives and head allocations, and
Fig.~\ref{fig:width} varies the constraint width $K$ for the capped
objective.

\begin{table}[!t]
  \centering
  \caption{Ablation (three-seed means; accuracy $\pm$ SD). $G_{\mathrm{exp}}$:
$G_{\mathrm{expert}}$ (\%); E-QC, Adj., P-QC: expert-referenced,
area-adjusted, prior-referenced QC (pp); --: not scored;
$^{\ast}$attention measures over two seeds.}
  \label{tab:ablation}
  \begingroup
  \setlength{\tabcolsep}{2.6pt}
  \renewcommand{\arraystretch}{1.06}
  \footnotesize
  \begin{tabular}{@{}lccccc@{}}
    \toprule
    Configuration & Acc. (\%) & $G_{\mathrm{exp}}$ & E-QC & Adj. & P-QC \\
    \midrule
    No spatial supervision & $84.23\pm0.46$ & 9.23 & $+0.88$ & $-0.04$ & $-0.06$ \\
    \midrule
    \multicolumn{6}{@{}l}{\emph{Objective, $K=3$, $\lambda=1/2$}} \\
    \quad Capped (ours) & $84.26\pm0.77$ & 10.89 & $-0.07$ & $-0.14$ & $+0.10$ \\
    \quad MSE & $84.04\pm0.88$ & 11.02 & $+1.12$ & $-0.09$ & $-0.28$ \\
    \quad Uncapped      & $83.47\pm0.80$ & -- & -- & -- & -- \\
    \multicolumn{6}{@{}l}{\emph{Objective, $K=8$}} \\
    \quad Capped, $\lambda=1/2$ & $84.29\pm0.33$ & 12.90 & $-1.67$ & $-0.44$ & $+0.95$ \\
    \quad Pooled, $\lambda=1/2$ & $83.47\pm0.69$ & 14.06 & $+0.87$ & $-0.42$ & $+0.38$ \\
    \quad MSE, $\lambda=2$ & $83.85\pm0.47$ & 14.59 & $+1.21$ & $-0.27$ & $-0.22$ \\
    \quad Uncapped, $\lambda=1/2$ & $82.81\pm0.71$ & -- & -- & -- & -- \\
    \multicolumn{6}{@{}l}{\emph{Head allocation, capped, $K=3$, $\lambda=1/2$}} \\
    \quad Knockout (ours) & \multicolumn{5}{c}{same as Capped, $K=3$} \\
    \quad Pearson$^{\ast}$ & $84.10\pm0.49$ & 10.47 & $-0.14$ & $-0.14$ & $+0.39$ \\
    \quad Random  & $84.20\pm0.93$ & 10.66 & $-0.30$ & $-0.09$ & $+0.31$ \\
    \bottomrule
  \end{tabular}
  \endgroup
\end{table}

Every prior-supervised model scored against the expert boxes
(Table~\ref{tab:ablation} and Fig.~\ref{fig:width}) places more attention
on the named structure than the model trained without spatial
supervision, raising $G_{\mathrm{expert}}$ from 9.23\% to between 9.55\%
and 14.59\%, and all of them keep accuracy within 0.8 percentage points
of that baseline. The automatic prior does move attention toward
annotated anatomy, even though no expert annotation is used in training.

How much it does so, and at what cost, is set by the constraint width.
As $K$ grows from 0 to 8 (Fig.~\ref{fig:width}(a)), $G_{\mathrm{expert}}$
rises at every step, from 9.23\% to 12.90\%, and Expert-QC falls at every
step, from $+0.88$ to $-1.67$~pp. Accuracy follows neither trend; it
stays between 83.98\% and 84.29\% for all widths
(Fig.~\ref{fig:width}(c)). This fits a prior that often fails to
distinguish the two questions of a pair or points to the wrong structure
(Section~\ref{sec:exp:independent}): as more heads are pulled toward it,
the question-dependent shift of attention agrees less with the
expert-defined contrast and, at $K=5$ and $K=8$, runs against it.
The prior-referenced measure instead rises, to $+0.95$~pp at $K=8$
(Fig.~\ref{fig:width}(b)), while the expert-referenced value is
$-1.67$~pp. Accuracy alone would not reveal this 2.6-pp gap.

The cost depends on the form of the objective as well as on its width. At
$K=3$, capped and MSE supervision place almost the same attention on the
named structure (10.89\% and 11.02\%), yet MSE supervision retains a
positive Expert-QC ($+1.12$~pp) while capped supervision does not
($-0.07$~pp). At $K=8$, the pooled and MSE objectives also stay positive
($+0.87$ and $+1.21$~pp), whereas the capped objective falls to
$-1.67$~pp. A plausible explanation is that the capped penalty vanishes
once each constrained head reaches the threshold, even when that mass lies
on regions shared by the priors of both questions, whereas MSE matching
penalizes any deviation from the full prior map. The area-adjusted measure, which normalizes each reference to
unit mass to reduce the influence of target size, is close to zero or negative for every model, including the one
trained without spatial supervision ($-0.04$~pp), so the raw values partly
reflect target size. The comparisons above still hold under it: capped
supervision lies below MSE supervision at $K=3$ ($-0.14$ versus
$-0.09$~pp), and the capped $K=8$ model is the lowest of all
($-0.44$~pp).

Which heads are constrained matters much less than how many. At $K=3$,
heads ranked by knockout, by Pearson correlation, or chosen at random give
accuracies of 84.26\%, 84.10\%, and 84.20\% and Expert-QC values of
$-0.07$, $-0.14$, and $-0.30$~pp, all within the seed-to-seed variation
(Table~\ref{tab:ablation}). Head ranking is a reproducible way to place
the constraint, not the source of its effect.

Attention on the named anatomy and its expert-aligned question
conditioning are thus distinct properties that need not move together.
Under capped supervision, widening the constraint raises the first and
lowers the second while answer accuracy changes little; MSE supervision
retains a positive raw Expert-QC at both widths, as does pooled
supervision at $K=8$, although their area-adjusted values are also
negative. At each evaluated width, capped models show lower
Expert-QC than the MSE and pooled models, alongside the more stable
accuracy of Section~\ref{sec:exp:objective}; narrow widths, including our
default $K=3$, keep this cost moderate while still raising the attention
on the named structure. We next examine how errors in the automatic prior
contribute to the disagreement between the two references.

\subsection{Independent Expert-Referenced Evaluation}
\label{sec:exp:independent}

Sections~\ref{sec:exp:objective} and~\ref{sec:exp:tradeoff} show how
much attention the objectives place on the prior and on the named
structure. The remaining questions are where the question-dependent part
of that attention comes from and whether it follows the anatomy named in
the question. If supervision transferred only correct spatial
information, evaluation against the prior and against independent expert
boxes should lead to the same conclusions. We test this by breaking the
question--region pairing of the prior and by scoring the same trained
models against both references.

\begin{table}[!t]
  \centering
  \caption{Pairing intervention. Prior-QC and anatomy mass (attention inside any
annotated structure) on 605 validation pairs from 93 images; test
accuracy $\pm$ SD over three seeds. $K=8$: an independent training run
whose test-pair QC (expert $-1.63$, prior $+0.87$~pp) reproduces
Table~\ref{tab:ablation} ($-1.67$, $+0.95$). A full-image prior has
zero capped loss, i.e., the objective of the first row.}
  \label{tab:pairing}
  \begingroup
  \setlength{\tabcolsep}{3pt}
  \renewcommand{\arraystretch}{1.08}
  \footnotesize
  \begin{tabular}{@{}lcccc@{}}
    \toprule
    Prior & $K$ & Accuracy (\%) & Anatomy (\%) & Prior-QC (pp) \\
    \midrule
    None ($\lambda=0$)   & 0 & $84.23\pm0.46$ & 12.87 & $-0.04$ \\
    Other image          & 3 & $84.42\pm0.67$ & 15.92 & $0.01$ \\
    Question-conditioned & 3 & $84.26\pm0.77$ & 14.80 & $0.19$ \\
    Other image          & 8 & $84.42\pm0.98$ & 19.53 & $0.02$ \\
    Question-conditioned & 8 & $83.85\pm0.30$ & 17.38 & $0.54$ \\
    \bottomrule
  \end{tabular}
  \endgroup
\end{table}

\begin{figure}[!t]
  \centering
  \includegraphics[width=\linewidth]{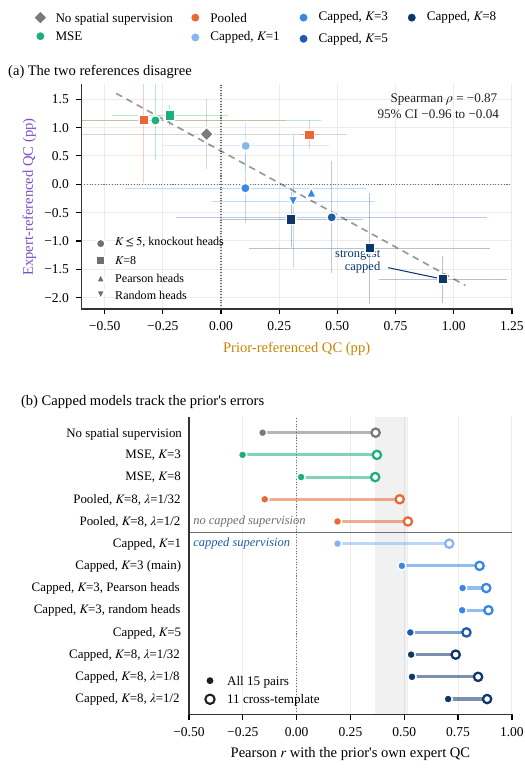}
  \caption{Prior- versus expert-referenced evaluation of the 13 configurations
with expert scores (models of Table~\ref{tab:ablation} and
Figs.~\ref{fig:objective}--\ref{fig:width}). (a)~QC under both
references (seed means $\pm$ SD; two seeds for Pearson-ranked heads;
dashed: least-squares fit). (b)~Pearson $r$ across pairs between each
model's Expert-QC and the prior's own: all 15 pairs (filled), 11
cross-template pairs (hollow); gray: cross-template range without
capped supervision.}
  \label{fig:reference}
\end{figure}

Breaking the pairing removes the prior-referenced response
(Table~\ref{tab:pairing}). When every training question receives the
prior of a different image, the response falls from 0.19 to 0.01~pp at
$K=3$ and from 0.54 to 0.02~pp at $K=8$ (difference 0.52~pp; 95\%
interval 0.24 to 0.84), while the attention on annotated anatomy rises.
The question-dependent response is learned from the question--region
pairing of the prior, and attention mass on anatomy alone cannot separate
a question-specific response from a generic one.

The pairing itself, however, is often wrong. On the 90 test targets, the
prior has a precision of 11.0\% against a chance level of 4.4\%, and a
recall of 66.7\%; in 4 of the 15 test pairs both questions receive the
same prior, and the prior's own expert-referenced QC is negative on
average ($-3.5$~pp). As a result, the two references rank the 13
evaluated configurations in largely opposite order
(Fig.~\ref{fig:reference}(a); Spearman $\rho=-0.87$, image-bootstrap 95\%
interval $-0.96$ to $-0.04$). The reversal is concentrated in the four
matched-template pairs, whose questions differ only in the named
structure ($\rho=-0.97$; $-0.26$ on the other 11 pairs). The strongly
constrained capped models score best against the prior and worst against
the expert boxes. The reversal also holds within a single training series: the capped
$K=8$ models of Table~\ref{tab:pairing}, trained alongside the MSE model
of Table~\ref{tab:ablation}, exceed it against the prior on the test
pairs ($+0.87$ versus $-0.22$~pp) and on 605 validation pairs
(difference 0.33~pp; 95\% interval 0.11 to 0.55), but trail it by
2.84~pp against the expert boxes ($-1.63$ versus $+1.21$~pp; 95\%
interval $-5.29$ to $-0.70$).

\begin{figure*}[!t]
  \centering
  \includegraphics[width=\textwidth]{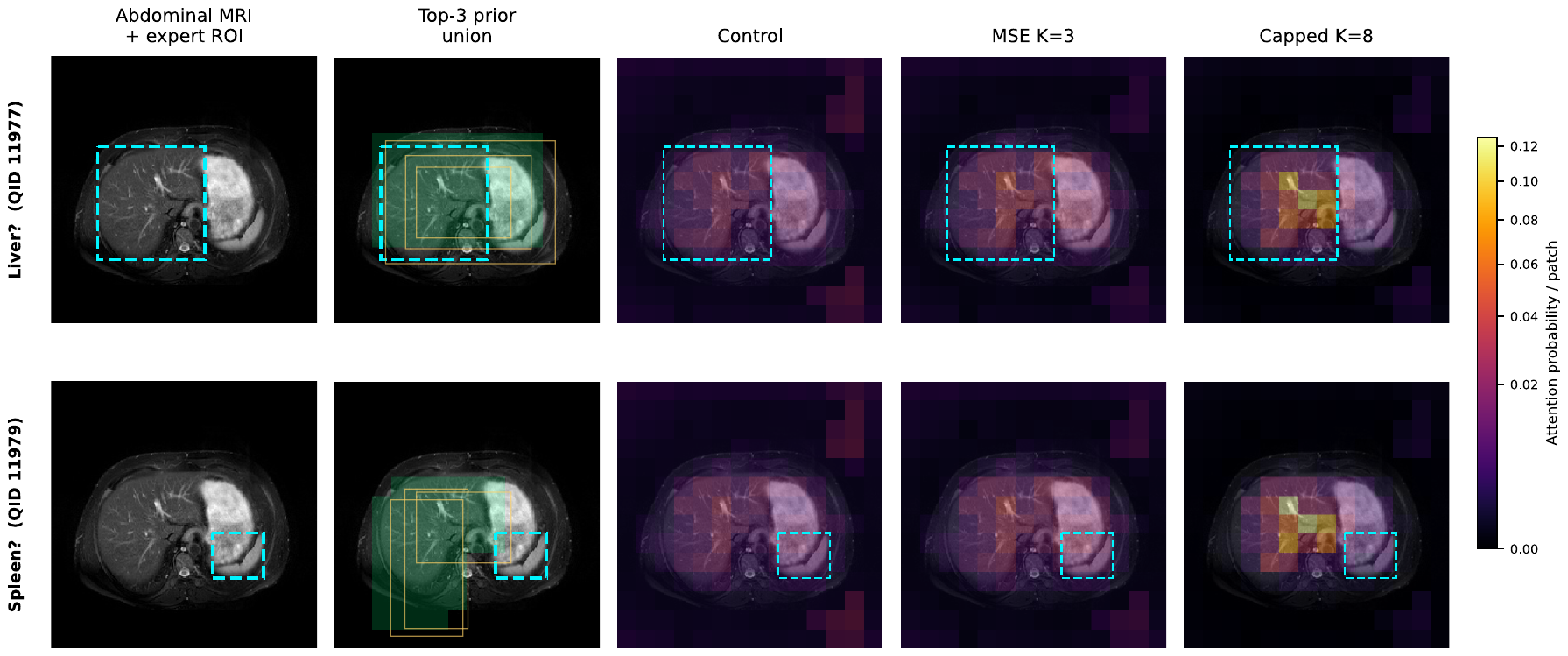}
  \caption{``Does the picture contain liver?'' (top) and ``\ldots spleen?''
(bottom) on one MRI slice: expert boxes (dashed), top-three prior union
(green), and head-averaged [CLS] attention without spatial supervision,
with MSE ($K=3$), and with capped supervision ($K=8$, $\lambda=1/2$;
models of Table~\ref{tab:ablation}); three-seed means on a shared
nonlinear color scale.}
  \label{fig:case}
\end{figure*}

Fig.~\ref{fig:reference}(b) links the reversal to the prior's errors.
Over the 11 cross-template pairs, every capped model follows the prior's
own expert-referenced QC more closely than any configuration without
capped supervision ($r=0.71$--$0.89$ versus $0.37$--$0.52$). Over all
pairs, the same holds from $K=3$ upward ($0.49$--$0.77$ versus
$-0.25$--$0.19$), and the correlation grows with $\lambda$ (from $0.53$ to $0.70$ at
$K=8$) and with $K$, from $0.19$ at
$K=1$ to $0.70$ at $K=8$.
Fig.~\ref{fig:case} shows one pair. The prior for the spleen question
lies 64\% on the liver box and 1.2\% on the spleen box. All models favor
the liver box, but capped supervision at $K=8$ amplifies this bias: asked
about the spleen, it places 85\% of its attention on the liver box and
less than the uniform level on the spleen box, whereas MSE supervision at
$K=3$ raises the spleen-box attention to 2.8 times the uniform level. On
this pair, the capped model has the highest prior-referenced QC of the
three ($+2.1$~pp) and the lowest expert-referenced QC ($-7.8$~pp).

Spatial supervision thus transfers the question-specific structure of
the automatic prior, including its errors, and capped supervision
transfers more of it as the constraint widens and tightens. A score
computed against the training prior rewards this transfer, so it cannot
substitute for an independent anatomical reference. These findings
concern attention as an observable response rather than an explanation of
the answer~\cite{jain2019attention, wiegreffe2019attention}, and they
rest on 15 question pairs from 13 images.

\section{Conclusion}
\label{sec:conclusion}

QCAP-VQA supervises cross-attention with an automatic,
question-conditioned prior through a capped prior-mass objective.
Unlike objectives that keep rewarding prior concentration, the cap
stops applying spatial pressure once each constrained head reaches a
mass threshold, while answer accuracy remains relatively stable
across the tested regularization strengths. Independent
expert-referenced evaluation reveals a limitation: under capped
supervision, attention on the named anatomy increases, but Expert-QC
declines and eventually becomes negative as more heads are constrained.
The pairing intervention and prior audit show that this decline is
associated with question-specific errors in the automatic prior, which
the supervision transfers to model attention. Prior-referenced scores
reward these errors, ranking models in largely opposite order to the
expert reference. These findings favor narrow constraint widths and
evaluation against references independent of the automatic prior. The
expert-referenced question-conditioning analysis is limited to 15
question pairs from 13 SLAKE images and concerns attention as an
observable response rather than an explanation of the answer; larger
annotated cohorts and more accurate priors are needed to assess the
generality of these findings.

\bibliographystyle{IEEEtran}
\bibliography{main}

\end{document}